\documentclass{article}

\usepackage{times}
\usepackage[T1]{fontenc}
\usepackage{amsmath}

\usepackage{vmargin}
\setmarginsrb{3.2cm}{3cm}{3.2cm}{3cm}{0.4cm}{0.8cm}{0.4cm}{1.5cm}

\begin{document}

\title{Verbal chunk extraction in French \\ using limited resources \thanks{Thanks to Zulema Solana for her comments on a previous version of the paper, the responsability of which is entirely ours.}}

\author{Gabriel G. Bès 
\and Lionel Lamadon \\ 
\and François Trouilleux}

\date{}

\maketitle

\vspace*{-2\baselineskip}
\begin{center}
Université Blaise-Pascal \\
34, avenue Carnot\\
F-63037 Clermont-Ferrand Cedex, France \\
Gabriel.Bes@univ-bpclermont.fr, Lionel.Lamadon@univ-bpclermont.fr, trouilleux@lrl.univ-bpclermont.fr

\vspace*{1\baselineskip}
August 26, 2004
\end{center}

\begin{abstract}
A way of extracting French verbal chunks, inflected and infinitive, is explored
and tested on effective corpus. Declarative morphological and local grammar rules
specifying chunks and some simple contextual structures are used, relying
on limited lexical information and some simple heuristic/statistic properties obtained
from restricted corpora. The specific goals, the architecture and the formalism
of the system, the linguistic information on which it relies and the obtained results
on effective corpus are presented.
\end{abstract}

\section{Introduction}

It is generally admitted that taggers or chunkers, either statistical
or grounded on linguistic knowledge, will never reach
perfect results~\footnote{Abney \cite{abney-96} gives the following example: ``to correctly disambiguate {\it help} in {\it give John help$_N$} versus {\it let John help$_V$}, one apparently needs to parse the sentences, making reference to the differing subcategorization frames of {\it give} and {\it let}.''}. However, it is also generally admitted that there are cases where, using only local contextual part-of-speech information and chiefly
information coming from morphological items, deductions on categories or chunks can be made with utmost certainty~\footnote{Tapanainen and Voutilainen \cite{tapanainen-94} discuss this point and give the following example: ``we can safely exclude a finite-verb reading if the previous word is an unambiguous determiner.''}. 

In this situation, the crucial question is: what results can be obtained with such and such information~? or, conversely, what information does one need to obtain this particular result ? We here try to address these issues with the presentation of a system that has as its immediate target the identification of verbal chunks --- finite and infinitive --- in French. 

We will readily agree that this target is fairly limited, compared to that accomplished by general purpose taggers and chunkers. However, to dig into a problem, common practice is to split it into smaller ones: our goal is to identify as clearly as possible the possibilities and limits of different types of information
with respect to specific structures, {\it i.e.}, for the matter, verbal 
chunks~\footnote{Our system belongs thus to partial parsing (cf. Manning \& Schütze 
\cite[p. 375]{manning-99}) but with verbal chunks as the target of the analysis and not
the most currently practiced nominal phrases. The reason of our choice is the potentiality
of inferences on the syntactic analysis of the whole sentence that can be driven from
analyzed verbal chunks; on the importance of correctly tagging verbs, see 
Chanod \& Tapanainen \cite{chanod-XX} and footnote 6.}.

Given this issue, the general and strong requirement which underlies the specification of our verbal chunk extractor is: do the more with the least. That is, we want to maximally reduce the resources necessary to put the system into use, either lexical information or tagged corpus, but keeping in mind that this must not hamper final results on effective texts.
Our general underlying hypothesis is : with (i) a simple linguistic modelization,
chiefly relying on morphological expressions in verbal, nominal and prepositional chunks,
plus (ii) very limited lexical resources, plus (iii) heuristic statistical information
on forms not covered by the previous (i) and (ii), it is possible to obtain results
standardly considered as good, {\it i.e.} around 98\% of recall and 
precision~\footnote{Manning \& Schütze \cite[p. 371]{manning-99} report the range of
95\% to 97\% for accuracy numbers calculated over all words; Charniak 
\cite{charniak-97} reports 97\%; Habert \& al. \cite{habert-al} recall the standard figures of
95\% to 98\%. Reported figures for French are 96.8\% for a statistical tagger
and 98.7\% for a constrained based one with heuristic rules 
(see Chanod \& Tapanainen \cite{chanod-XX}). In general it is difficult to state
clearly revealing figures given the different parameters intervening in the
tagging tasks.}.

As a consequence, the system is centrally rule-based, but is statistically 
completed. The satisfaction of the general requirement on resource poverty settles the balance between the two types of contributions. We use as much as possible a simple linguistic modelization, which exempts from the use of large tagged corpora, and we complete it with very simple statistical information. People more or less agree on the respective advantages and disadvantages of corpus statistics vs. linguistic 
knowledge (see Chanod \& Tapanainen \cite{chanod-XX} and Bril \cite{bril-95} which
represent opposite orientations) and both approaches can or must -- ordinarily, they are --
be put to work together, the most important difference coming on the weight given to each one.

In general terms our system can be characterized as working in the opposite way
of the transfomational-based learning way of tagging (on this, see Bril \cite{bril-95}, 
Charniak \cite{charniak-97}, 
Manning \& Schütze \cite{manning-99})~\footnote{The system is thus, in {\it spirit} if not necessarily in expressive or covering power, in the line of 
Chanod \& Tapanainen \cite{chanod-XX}, 
the EngCG tagger \cite{voutilainen-97}, Tapanainen \& Voutilainen \cite{tapanainen-94},
Tapanainen \& Järvinen \cite{tapanainen-xx}.}. 

It is known that the transfomational way of tagging is organized in two steps. In the
first step, a simple algorithm records for each word its most common part of speech 
in the training corpus. The tagging of new texts consists in simply assigning
its most common tag to each word. The reported results of this level are 90\%
accuracy. In the second step, rules are proposed and evaluated in order to change
the tags obtained in the first step into others intended to be more accurate.
E.g. related to our issue, a rule can be:

\begin{quote}
{\it Change a verbal tag into a nominal tag if the previous word is a non ambiguous determiner.}
\end{quote}

Transformations are not specified in terms of linguistic knowledge. Given 
$\mid$PoS$\mid$ the total number of parts of speech, there are $\mid$PoS$\mid^3$
possible transformational rules of the previous type. The necessary training data - {\it i.e.}
tagged corpus of several hundred thousands words - are then put into use. The system
tries each possible rule on the training data measuring the accuracy the rule induces.
And Charniak reports:

\begin{quote}
Some rules make things worse, but many make things better, and we pick the rule that
makes the accuracy the highest. Call this rule~1.

Next we ask, given we have already applied rule 1, which of the other rules does the 
best job of correcting the remaining mistakes. This is rule 2. We keep doing this
until the rules have little effect. The result is an ordered list of rules,
typically numbering 200 or so, to apply to new examples we want tagged.
\end{quote}

In our system, structural rules (see below section 5) express the simple needed
linguistic modelization. They are the counterparts to the transformational
rules for tags changing, but they are not learned by some algorithm and they are not
used for erasing or modifying already obtained results coming from a previous step.
Our heuristic/statistical information is the analogous of statistical information
calculated on the first step of the transfomational-based learning strategy, with the big differences that it does not concern all words related to verbal chunks, but only
the ones whose tags are not assigned by structural rules, and that our ``training'' corpus is
extraordinary reduced compared with training corpora used in the statistical ways of doing things.

In the following, section \ref{task} specifies the targeted task of the system, section \ref{system} presents the system architecture, section \ref{resources} describes linguistic resources and section \ref{rules} illustrates rules of the system.
Section~\ref{results} is devoted to the analysis of results obtained by a detailed evaluation: we are not only interested by the obtention of global figures 
but also by a more detailed one, allowing to discriminate between the different types of parameters which contribute to global results. Particularly, we want to discriminate 
between results obtained thanks to the linguistic modelization and results obtained
thanks to heuristic/statistical choices, to discriminate between missing resources and effective counter-examples, and we want also to be sure that ``good'' results do not 
necessarily come from a high or dominant percent of non ambiguous 
forms~\footnote{Manning \& Schütze \cite[p.~374--375]{manning-99} points that ``some authors
give accuracy for ambiguous words only in which the accuracy figures are of course
below''. In the literature reporting results it is not always possible to find discriminated
results for ambiguous and not ambiguous forms. Even in Chanod \& Tapanainen \cite{chanod-XX}
where actually some figures are discriminated, the ones on the overall performance
of the reported system are not. The point is important from a methodological viewpoint;
see Habert \& al. \cite{habert-al} which points futhermore that tagging errors have quite different
consequences on the syntactic analysis of the whole sentence depending on the
kind of ambiguity: a tagging error on the noun/verb ambiguity will more significantly hamper the results on the syntactic sentence structure than a tagging error on the 
past participle/adjective ambiguity.}.
The perspectives of the system are discussed in the final section \ref{perspectives}.

\section{Task Definition}
\label{task}

In French, verbal chunks may include, in addition to a verb nucleus:
\begin{itemize} 
\item auxiliary verbs, of two types as in {\it Jean \underline{a mangé}} or in {\it Jean \underline{est reparti}},
\item clitic pronouns, as in {\it Jean \underline{le lui prend}},
\item negation particles, as in {\it Jean \underline{ne parle pas}},
\item adverbs, as in {\it Jean \underline{a rapidement mangé}},
\item occasionally, incidental phrases or clauses, as in {\it Jean \underline{a, par ailleurs, mangé}},
\item and, in infinitive verbal chunks, a preposition, as in {\it Il parle \underline{de le prendre}}.
\end{itemize}

With the only exception of the negative particle {\it ne}, and of some reflexive forms, all the morpheme 
expressions in verbal chunks are ambiguous :  {\it le} can be an article or a clitic 
pronoun, {\it lui} a nominative or a dative clitic pronoun, {\it en} a preposition or a clitic 
pronoun, {\it pas} a particle in negation constructions or a noun, etc. Futhermore, 
many verbal expressions are ambiguous: {\it note} can be a verb or a noun, {\it avoir} 
a noun or the infinitive of one of the two auxiliary forms, {\it savoir} a noun or an 
infinitive, {\it entre} a verb or a preposition, {\it inverse} a verb or an adjective, {\it lui} 
the past participle of the verb {\it luire} or the clitic or the nominative pronoun, 
{\it soit} a verb or a conjonction, etc. Besides this, verbal forms used as auxiliaries
can also be used as main verbs as in {\it Jean est triste} ({\it John is sad}) ou {\it Marie a la migraine} ({\it Mary has got a headache}).
But, despite this superabundance of several different kinds of ambiguities, for a human reader,
effective texts are never ambiguous with respect to verbal chunks.

One central challenge of verbal chunk extraction is thus to solve morphological and/or lexical and/or functional ambiguities. But we do
not think that disambiguation of items in the
whole sentence or, even,  in the targetted structures, is a step 
which must necessarily be achieved before the obtention of chunks. In any case, we want to find out how far one can go with different kinds of limited information and combining disambiguation with chunk 
extraction~\footnote{Manning \& Schütze \cite[p.~374-375]{manning-99}, observes that
although the interest in tagging comes from the belief that many engineering
applications to language processing must benefit from syntactically disambiguated
texts, ``it is surprising that there seem to be more papers on stand-alone tagging
than to applying tagging to a task of immediate interest''. Our system
aims at overcoming this criticism by combining tagging with chunk extraction.}.

\section{System Architecture}
\label{system}

The system consists in two modules: Smorph,  which performs tokenization and morphological analysis (Aït-Mokhtar \cite{ait-98}), and Pasmo (Paulo et al. ~\cite{paulo-01}), which, given the output of Smorph, delimits sentences and, in the present case, verbal chunks, based on ``recomposition rules'' similar to rewriting rules. Smorph and Pasmo are related by a script interface tagging unknown
words by Smorph. In addition, the system is associated to an evaluation software (Trouilleux \cite{trouilleux-04}) which can provide detailed evaluation results (cf. section \ref{results})~\footnote{Thanks to Mourad Sahi for his contribution with a program to compile results obtained on a set of files.}.

Given a text file, Smorph outputs a list of tokens of the form:
\begin{quote}
     {\tt surface\_form  [lemma, FVL]}
\end{quote}
where {\tt FVL} is a feature/value list.

A Pasmo recomposition rule is of the form:
\begin{quote}
     {\tt token$_1$ ... token$_n$ -> token$_{1'}$ ... token$_{n'}$}
\end{quote}
with the reading ``a sequence of tokens matching the left-hand side of the rule is to be recomposed into the sequence of tokens specified in the right-hand side of the rule'' ({\it i.e.} basically, the reading is the inverse of that of traditional rewriting rules). The Kleene + operator may be used in the left-hand side of the rule. In the new tokens produced by a rule, the {\tt surface\_form} and {\tt lemma} elements may either be specified as specific strings of characters or as the concatenation of the {\tt surface\_form} and {\tt lemma} elements of the input tokens, while the feature/value pairs are to be specified as strings of characters only. Examples of rules are given in section \ref{rules}.

The data structure output by application of Pasmo rules is the same as that output by Smorph, allowing cyclic application: the rules are first applied to the output of Smorph, producing output 1, then applied again using output 1 as an input, producing output 2, then applied again, \ldots until the input is such that no rule applies.

The purpose of this paper is not so much to introduce a new chunking system, but rather to analyse the importance of different types of information in a chunking process
related to verbal forms. We thus won't go into a detailled comparison of Smorph-Pasmo with existing systems. 

\section{Lexical  Linguistic Information}
\label{resources}

The declarative source of Smorph is a lexicon in which the linguist declares lemmas and the set of feature/value pairs associated to each of them in terms of their inflected forms.
It is possible to express in Smorph compound words, hyphenated, 
as in {\it remue-ménage} ({\it commotion}) or
not, as in {\it pomme de terre} ({\it potatoe}). 

Contrary to general practice, the basic choices with respect to our lexicon are:
\begin{itemize}
\item With the exception of verbal forms which are ambiguous as {\tt fl} (inflected) or {\tt pp} (past participle), ambiguity is not expressed by the multiplication of the same forms. E.g. there are not two forms {\it le}, one associated
to a value {\tt art[icle]} and the other to a value {\tt cl[itic]}, nor two forms {\it note},
one associated to {\tt n[oun]} and the other to {\tt v[erb]}; instead there is a unique form
{\it le} associated to the value {\tt ocl} of the feature {\tt AMB[iguity]}, which expresses the potential ambiguity of the clitic, and a unique form {\it note} associated to the value 
{\tt ov} of the same feature, which expresses the potential ambiguity of the verb.
\item With the only exceptions of plural first and second person verb forms, no flexional values are associated to morphological or lexical forms~\footnote{The value is needed in order to distinguish
contextually a subject {\it nous}, as in {\it nous voulons} ({\it we want}) from a clitic one, as in {\it Il nous parle} ({\it He talks to us}).}.
\item Lexical categories are drastically reduced. There is no noun category, and adjective entries are less than ten.
\end{itemize}

In its present state, as far as expressions occurring inside verbal chunks are concerned, our lexicon contains the following forms: 
\begin{enumerate}
\item all the verbal forms of our ``training'' and test corpus~\footnote{In order to examine the minimal lexical information (in terms of categories) required for our proposed task, before running the system on a given corpus, we make sure that to all the verbal and potentially verbal forms in the corpus correspond an appropriate entry in the lexicon. In other words, lexical coverage is not evaluated here. We intend to evaluate hypotheses, not resources.},
\item all the morphological forms which can appear in verbal chunks: clitic pronouns, {\it ne}, {\it pas}, prepositions in infinitive chunks ({\it à, de}, etc.),
\item a subset of a restricted class of adverbs as {\it ainsi}, {\it d'ailleurs}, {\it alors},
\item a subset of multiword frozen adverbial expressions which incorporate forms which, in isolation, can be verbs (e.g. {\it sans doute, n'importe qui}, where {\it doute} and {\it importe} alone can be verbs).
\end{enumerate}
Items (1) and (2) are intended to be exhaustive, while this is not the case of
items (3) and (4). Verbal forms are categorized as follows:
\begin{itemize}
\item inflected forms {\tt fl} are either categorized as {\tt ov} 
(potentially ambiguous forms), as {\it note},  or categorized {\tt nv} (not potentially ambiguous forms), as {\it accepte}; infinitive forms {\tt inf}
are either categorized as {\tt oin} (potentially ambiguous forms), as {\it devenir},
 or categorized {\tt nin} (not potentially ambiguous forms), as {\it assurer},
\item a subset of the potentially ambiguous inflected forms (category {\tt ov}) is further categorized as having a statistically unlikely verb reading, (category {\tt jv}, {\tt j[amais] v[verbe]}, {\it never verb}), as {\it empire},
\item past participles, whether they are ambiguous with respect to nouns (e.g. {\it entrée}) or not (e.g. {\it dormi}), belong to a unique category {\tt pp},
\item verbal forms which are ambiguous as {\tt fl} (inflected) or {\tt pp} (past participle),
e.g. {\it fait}, are associated in the lexicon to two different forms, one categorized
as {\tt fl} and the other as {\tt pp}. 
\end{itemize}

In addition to this categorization, as stated above, first and second person verbal forms (and only those) are associated to flexional information in the form of a person and number feature. 

There are in all in the lexicon 1.731 verbal forms: 658 {\tt fl}, 595 {\tt pp}, 478 {\tt inf}.
With respect to ambiguity, the significant figures are: 429 (about 2/3) {\tt fl} forms are {\tt nv} while 229 are {\tt ov}; 440 (92\%) {\tt inf} forms are {\tt nin} while 38 
are {\tt oin}~\footnote{There were no examples of error expressions in the results concerning
ambiguous {\tt pp} forms.}.

In addition to forms occurring in verbal chunks, the lexicon also contains some prepositions, determiners which are not ambiguous with forms inside verbal chunks, as {\it un, des, ce...}, unambiguous possessives, as {\it mon, ses...}, a small set of adjectives
(less than ten), a small set of pronouns, as {\it il, on...} and ponctuation marks.
The forms other than ponctuation marks which are not verb forms and which can appear
outside verbal chunks (this is not the case, e.g. of nominative or some
reflexive clitics as {\it -il, me, se}), are all categorized as  
{\tt auf} ({\tt au[tre] f[orme]}, {\it other form}). There are in all 219 {\tt auf}
forms in the lexicon.

With so limited lexical resources it is obvious that not all token forms
in texts will be associated by Smorph to some feature/value categorization. In fact, in both 
the ``training'' and test corpus a little more than 1/3 of the forms are not
recognized by Smorph and are associated by the script interface to [{\tt au, auf}]
feature values ({\tt au} being a feature value not used in the lexicon). Overall, 2/3 of the tokens 
are associated to the most frequent
feature value  {\tt auf}.

\newpage
\vfill

The following is Smorph output for {\it Il la note} 

\begin{verbatim}
'Il'. 
[ 'il', 'TPRO','pnom', 'TPASS','auf']. 

'la'. 
[ 'la', 'TFG','cl', 'AMB','ocl', 'TPASS','auf']. 

'note'. 
[ 'noter', 'TFV','v', 'MOD','fl', 'AMB','ov']. 
\end{verbatim}

\noindent where, besides the already commented values {\tt auf}, {\tt fl}, {\tt ov}, the values {\tt pnom}, {\tt cl}, {\tt v}, {\tt ocl} stand for {\it nominative pronoun, clitic, verb, ambiguous clitic}, respectively.

\section{Rules}
\label{rules}

Four basic types of rules are expressed in the same overall formalism:
\begin{enumerate}
\item	Structural disambiguation rules of potentially ambiguous verbal forms ({\tt ov}) which are actually not in chunks.

\item	Rules for the construction of verbal chunks
	\begin{enumerate}
	\item not potentially ambiguous,
	\item potentially ambiguous.
	\end{enumerate}

\item	Rules for the disambiguation of potentially ambiguous chunks.

\item	Rules for the disambiguation of forms not disambiguated by (1) to (3).
\end{enumerate}

The rules are illustrated in the following in a semi-formal notation, where
capital letters stand for variables and feature/value pairs are reduced to a minimum, {\it i.e.} a value. Table \ref{cat} gives the key to the symbols used in the examples.

\begin{table}[b]
\begin{center}
\begin{tabular}{|l|l|p{1cm}|l|l|}
\cline{1-2}\cline{4-5}
\multicolumn{2}{|l|}{\it Lexical forms}  && \multicolumn{2}{|l|}{\it Chunks} \\
\cline{1-2}\cline{4-5}
{\tt cl}   & clitic            && {\tt vnfl}  & not ambiguous verbal inflected chunk \\
\cline{1-2}\cline{4-5}
{\tt ocl}  & ambiguous clitic  && {\tt vnfla} & ambiguous verbal inflected chunk  \\ 
\cline{1-2}\cline{4-5}
{\tt v}    & verb              & \multicolumn{3}{l}{} \\
\cline{1-2}
{\tt ov}   & ambiguous verb    & \multicolumn{3}{l}{} \\
\cline{1-2}\cline{4-5}
{\tt nv}   & not ambiguous verb&& \multicolumn{2}{|l|}{\it Other symbols} \\
\cline{1-2}\cline{4-5}
{\tt det}  & determiner        && {\tt not-} & negation prefix  \\
\cline{1-2}\cline{4-5}
{\tt prep} & preposition       &&  {\tt +}    & concatenation operator \\
\cline{1-2}\cline{4-5}
{\tt pnom} & nominative pronoun&  \multicolumn{3}{l}{} \\
\cline{1-2}
\end{tabular}
\end{center}
\caption{\label{cat} Key to symbols used in the example rules.}
\end{table}

\enlargethispage{2\baselineskip}
Rules (1) concentrate on prepositional and nominal phrases, and on the initial sentence position. 

As an example, the rule
\begin{quote}
	\verb_X[det] Y[ov] --> X[det] Y[not-v]_
\end{quote}
specifies that an ambiguous verbal form following a determiner is not a verb
(e.g. {\it ce juge}).

Rules (1) are not intended to be exhaustive, but have been reduced to what was considered
to be a strict and sufficient minimum in order to obtain the proposed target.
Thus, it is apparent that many potentially ambiguous verb forms ({\tt ov})
remain: their potential ambiguity is not resolved.

Rules (2a) are specified around non ambiguous verbs ({\tt nv}) and also around potentially ambiguous ones ({\tt ov}), inasmuch these are contextually
disambiguated internally in the verbal chunk by the presence of an unambigous form (e.g. {\it ne}). 

As an example, the rule
\begin{quote}
	\verb_X[cl] Y[cl] Z[v] --> X+Y+Z[vnfl]_
\end{quote}
disambiguates {\it le}, {\it lui} and {\it donne} and constructs the chunk in 
{\it Jacques [le lui donne]$_\text{\tt vnfl}$.}

Rules (2b) specify potentially ambiguous verbal chunks containing at least two forms. This situation typically arises when chunks are formed by a clitic pronoun and a verb, the two being ambiguous.

As an example, the rule
\begin{quote}
	\verb_X[ocl] Y[ov] --> X+Y[vnfla]_
\end{quote}
specifies a  provisional ambiguous chunk for {\it la note} or {\it en charge}.

A challenge for rules (2) is the identification of incidental phrases. Two basic types of incidental phrases were foreseen, one between commas and the
other without, as in
\begin{itemize}
\item {\it Il a, très souvent, été invité.}
\item {\it Il a très souvent été invité.}
\end{itemize}
They are respectively described by the two following structures making use of
the Kleene + operator
\begin{quote}
\verb_, auf+ ,_ \\
\verb_auf+_
\end{quote}

These straightforwardly say : incidental phrases in verbal chunks are constructed by one
or more {\tt auf} expressions, which can be preceded and followed by commas. 
Depending on the verbal chunk, the point of insertion
of the incidental phrase can be one or another. Following the general strategy, 
not all possible incidental phrases were defined in our rules, but the general 
hypothesis is that the same structures are candidates to express all of them.

Rules (2a) and (2b) are intended to be exhaustive within their respective
scopes : Rules (2a) must build
all non ambiguous verbal chunks and Rules (2b) all ambiguous verbal ones with two forms. After application of these rules, some verbal forms (the {\tt ov} ones)
and verb chunks (the ones builded by Rules (2b)) remain thus to be disambiguated. This is the job of Rules (3) and Rules (4).

Rules (3) take advantage of the cyclic application of rules. After cycle I, the
already specified {\tt vnfl} can be used to disambiguate already specified {\tt vnfla} 
and {\tt ov} verb forms..

As examples, the rules
\begin{quote}
	\verb_X[vnfl] Y[vnfla] --> X[vnfl] Y[not-vnfl]_

	\verb_X[vnfl] Y[ov] --> X[vnfl] Y[not-vnfl]_
\end{quote}
will respectively be in charge of the resolution of {\it en charge} in 
{\it Il [n'est pas]$_\text{\tt vnfl}$ en charge d'une solution}, where {\it en charge} is not a verbal chunk, and of {\it compte} in {\it Il [tient]$_\text{\tt vnfl}$ compte d'une solution}, where {\it compte} is not a verbal form. By the same token but with opposite effects, the rules
\begin{quote}
	\verb_X[pnom] Y[vnfla] --> X[pnom] Y[vnfl]_
 
	\verb_X[pnom] Y[ov] --> X[pnom] Y[vnfl]_ 
\end{quote}
will respectively assign {\tt vnfl} to {\it la juge} in {\it Il la juge} and to 
{\it compte } in {\it Il compte venir}.

\pagebreak

It is at this point minimal statistics proceeds. Intuitively or using
very simple heuristic techniques on very restricted texts, it is easy to know
that,
given the disambiguation performed by rules (1),
 ambiguous verbal chunks containing at least two forms ({\tt vnfla}) are more frequent as nominal phrases than as verbal chunks (in particular as {\it le, la, les} are most likely to be determiners, and as {\it en} is more likely to be a preposition),
and, conversely, that ambiguous verbal forms ({\tt ov}) alone, excepted the {\tt jv} ones, are more frequent as verbal chunks than 
as nouns. Rules (4) express this.

We operate thus with rules that can be called {\it structural}, {\it i.e.} 
Rules (1) to (3), and with remaining Rules (4), which are poorly 
heuristically 
motivated rules solving the ``other'' cases . Our system tags as {\tt vnfl-I} 
and {\tt vninf-I} 
inflected and infinitive verbal chunks obtained by structural rules, 
and 
as {\tt vnfl-II} and {\tt vninf-II} inflected and infinitive verbal chunks 
obtained 
by the remaining rules.

\section{Results}
\label{results}

Our general underlying hypothesis can be now precisely stated: 
limited lexical resources as characterized in Section 4, plus the linguistically motivated structural rules (Rules (1) to (3)), plus simple statistical/heuristic 
rules (Rules (4)) can reach results in the targetted level ({\it i.e.} around 98\% recall and precision). Effectively obtained results, which, 
after the explicitation of the background for analysing them, 
will be presented below, do not falsify the hypothesis.

The obtained results must be analyzed in terms of the goals of the system on
one hand, and in terms of the actual kinds of implemented and/or foreseen information
on the other hand. We want to distinguish {\it missing resources,} which are foreseen 
but not implemented, from effective counter-examples to our hypothesis. E.g. if
in a corpus there is an adverbial frozen expression or an incidental clause
in a verbal chunk, foreseen but not implemented, we do not consider the errors resulting from such non-implementation as 
counter-examples to the hypothesis, but as missing resources.

We know, with respect to forms inside verbal chunks, that verbal
forms and morphological items are exhaustively implemented, while adverbs
and adverbial frozen expressions are not (see section 4). We know futhermore
that Rules (1) are not intended to be exhaustive, while
 Rules (2a) and (2b) are intended to be so, even if not all
formally foreseen incidental phrases are effectively implemented. We know finally
that a subset of ambiguous verbal ({\tt ov}) forms were classified on intuitive heuristic
grounds as {\tt jv} (see section 4), and that the {\tt vnfla} and {\tt ov} forms,
remaining as such after application of structural rules, are classified by Rules (4)
as {\tt non-v} forms and {\tt vnfl} forms, respectively (see section 5).

From these, it is possible to classify counter-examples as {\it structural}, {\it i.e.} 
related to structural rules and {\it heuristic}, {\it i.e.} related to Rules (4). Structural
counter-examples are further classified as being produced 
\begin{itemize}
\item[(i)] by incompleteness of Rules (1), 
\item[(ii)] by minor inaccuracies in the formulation of Rules (2), 
\item[(iii)] by lack of the expressive power of Rules (2).
\end{itemize}

The limits of the system and its underlying general hypothesis are given by
structural counter-examples produced by lack of expressive power of Rules (2)
together with heuristic counter-examples. Intuitively, the system reaches
its limits when it is not possible to modify it without modifying its general
assumptions.

To evaluate the system, we will use the classic recall and precision measures. In addition to these measures, we define {\it SL} (System Limits) as :

\[ SL = (1 - EE) * 100 \]

\noindent where $EE$ is the ratio of 'error expressions' in a corpus
test over the number of chunks actually observed in the corpus ({\it i.e.} the
denominator of the recall measure). An error expression is either 
an expression observed in the corpus test to which correspond one or more recall or precision errors~\footnote{\label{note_exemple}In the sentence, {\it Aucun groupe étranger n'a pour le moment fait connaître son intérêt}, one observes the finite verbal chunk {\it n'a pour le moment fait}. Given this sentence, the system actually identifies two finite verbal chunks {\it n'a} and {\it fait}. We here have one recall error (the observed chunk is not identified) and two precision errors (the system identifies two chunks which are not observed as such in corpus), but only one error expression: {\it n'a pour le moment fait}. The recall and precision errors are correlated and this is taken into account in the computation of $SL$.
Note that $SL$ ranges from 100 (perfect result) to $-\infty$. Negative values are obtained when the system produces more error expressions than there are observed expressions in the corpus.} or an expression output by the system as a verbal chunk which do not correspond in part or in totality to an actually observed verbal chunk.

Given this background, it is now possible to report and analyse effective results of our system.

We used a 13,000 word corpus of finance news articles to tune our system during development. This corpus will be referred to as the ``training'' corpus. We emphasize that the ``training'' corpus is not a previously tagged one. It is used neither to extract statistical information on n-grams with $n \geq 2$ nor to learn rules from it. Structural rules and the  lexical information 
which were needed were foreseen as working hypotheses. The contribution of 
the so-called ``training'' corpus is thus limited to the tuning of the system, to the
verification of heuristic choices related to Rules (4) inasmuch they concern
{\tt ov} and {\tt jv} forms, and to the detection of adverbs and
adverbial frozen expressions incorporated to verbal chunks. See in Table 2 the results
on the ``training'' corpus.

The system underlying the results of the ``training'' corpus was applied to
a 10,400 word, previously unseen, ``test'' corpus of the same source. After this application, we detected 13 adverbial expressions which were missing in incidental clauses and added to the lexical resources.  We refer to the state of the system at this stage as ``state 1''. The system in state 1 was applied to both the ``training''  and the test corpora.
Results are stated in Table 2. 
The {\it possible} and {\it actual} columns give the number of verbal chunks in the manually tagged corpus and in the output of the system, respectively.

\begin{table}[t]
\begin{center}
\begin{tabular}{|l|l|c|c|c|c|c|}
\hline
&     & {\it possible} & {\it actual} & {\it correct} & {\it recall} & {\it precision} \\
\hline
\hline
         & infinitive & 392  & 392   & 392   & 100.00  & 100.00  \\
\cline{2-7}
training & finite     & 819  & 822   & 814   & 99.39   & 99.03  \\
\cline{2-7}
         & all        & 1211 &  1214 &  1206 &  99.59  &  99.34  \\
\hline
\hline
                & infinitive & 285  & 283   & 282   &  98.95  &  99.65  \\
\cline{2-7}
test state 1   & finite     & 668  & 677   & 651   & 97.46   & 96.16  \\
\cline{2-7}
                & all        & 953  &  960  &  933  &  97.90  &  97.18  \\
\hline
\end{tabular}
\end{center}
\caption{\label{state1} Results on training and test corpus in state 1.}
\end{table}

\begin{table}[t]
\begin{center}
\begin{tabular}{|l|l|c|c|c|c|c|}
\hline
&     & {\it possible} & {\it actual} & {\it correct} & {\it recall} & {\it precision} \\
\hline
                & infinitive & 285  & 284   & 284   &  99.65  &  100.00 \\
\cline{2-7}
test state 2   & finite     & 668  & 673   & 661   & 98.95   & 98.22  \\
\cline{2-7}
                & all        & 953  &  957  &  945  &  99.16  &  98.75  \\
\hline
\end{tabular}
\end{center}
\caption{\label{state2} Results on test corpus in state 2.}
\end{table}

\begin{table}[t]
\begin{center}
\begin{tabular}{|l|c|c|c|}
\hline
{\it label} & {\it correct} & {\it actual} & {\it precision} \\
\hline
{\tt vninf-I}  & 273  & 273 & 100.00  \\
\hline
{\tt vninf-II} & 11   & 11  & 100.00  \\
\hline
{\tt vnfl-I}   & 615  & 618 & 99.51   \\
\hline
{\tt vnfl-II}  & 46   & 55  & 83.64   \\
\hline
\end{tabular}
\end{center} 
\caption{\label{regles} Precision figures for different rule types.}
\end{table} 

Analysis of results on the test corpus reveals two missing (but formally foreseen)
Rules (2) for inflected verbal chunks with incidental clauses, and four minor
inaccuracies in the formulation of Rules (2) (two related to inflected chunks and
two to infinitive ones).
The two missing rules for inflected verbal chunks were added and the four
 verbal chunks rules modified. The system in State~2 was thus obtained.
These modifications allowed the correct identification of 10 more finite verbal chunks 
and of 2 more infinitive verbal chunks (see Table \ref{state2}). On the ``training'' corpus, the system in state 2 produces the same results as in state 1, except for one additional error.

The analysis of error expressions
in the results of the application of the system in state 2 to the test corpus
allows to obtain a first estimation of $SL$ and its underlying general hypothesis.

On the test corpus, 16 error expressions remain. They are associated to 20 
recall or precision errors obtained by the system : 

\begin{itemize}
\item 2 error expressions concern both recall and precision (i.e. 4 overall errors), 
\item 1 error expression~\footnote{The example in footnote \ref{note_exemple} page \pageref{note_exemple}.} concerns 1 recall error and 2 precision errors (i.e. 
3 overall errors), 
\item 8 error expressions each concern a precision error only,
\item and 5 error expressions each concern a recall error only.
\end{itemize}

Out of 8 error expressions concerning precision errors, 3 arise from incompleteness of Rules 1. They affect finite verbal chunks but can be easily completed within the expressive power of the system. Thus they can be considered as counter-examples not affecting SL.

All of the other error expressions ({\it i.e.} 13) lead to  mistaken results which
are truly counter-examples counting for the calculation of SL. 

The 3 error expressions affecting both precision and recall are
produced by lack of expressivity of Rules (2), and as such, they are true counter-examples. They all involve finite verbal chunks with incidental clauses which are not marked by commas as in {\it L'encours \ldots [a pour sa part plus que doublé]}. The system cannot currently deal with such cases, because {\it part}, being a potential verbal form, is not an {\tt auf} (see section 4). We know that these error expressions
generate 7 recall or precision errors, all of them on finite verbal chunks.

The 10 remaining error expressions which are true counter-examples are all produced by statistical/heuristic failures. Only 1 of them is an infinitive verbal chunk (an error affecting recall). 4 affect the recall on finite verbal chunks and 5 the precision.
 
Thus we have the overall ratio of counter-example error expressions 
$EE \simeq 0.014\%$ ({\it i.e.} 13/953), which gives $SL\simeq 98.6\%$.

It is also interesting to remark that in the Smorph output of the test corpus
there are 464 potentially ambiguous inflected verbal forms {\tt ov} 
over 673 effective verb inflected chunks (69\%), 
and 38 potentially ambiguous infinitive verbal forms {\tt oin}
over 284 effective verb infinitive chunks (13.4\%). This is important because 
the challenge for disambiguation arises from {\tt ov} and {\tt oin} forms.
In the test corpus, figures of {\tt ov} tokens over {\tt vnfl} forms 
({\it i.e.} 464/673) practically inverse the ones in the lexicon of {\tt nv}
over {\tt fl} forms ({\it i.e.} 429/659) (see section 4).

The error expressions which are true heuristic counter-examples
can thus be reported to the challenge forms, {\it i.e.} 9/464 for
{\tt ov} forms, and 1/34, for {\tt oin} forms, which gives
$\simeq 1.9\%$ and $\simeq 2.6\%$ respectively. We claim thus
that the encouraging {\it SL} obtained does not come
from the fact that the challenging forms are weekly represented
in the test corpus.

The verbal chunks obtained by the structural rules (1) to (3) are 
tagged {\tt vnfl-I} and {\tt vninf-I}: it means that these chunks are defined  
structurally as verbal chunks. The verbal chunks obtained by the rules 
(4) are tagged {\tt vnfl-II} and {\tt vninf-II}. These chunks are heuristically
defined as verbal chunks. The glass box evaluation 
allows us to discriminate between the two sorts of tags and it is 
interesting to note that there is no significant difference between the 
precision obtained for both tags related to {\tt vninf}. Table \ref{regles} reproduces the results obtained for these tags on the 
test corpus.

\section{Perspectives}
\label{perspectives}

Neither all finite verbal chunks nor infinitive verbal chunks can be obtained by our system. However, it allows at this stage to reach a good level in the disambiguation task. Ongoing work deals with the evaluation of the cost, in terms of lexical resources and linguistic rules, to increase the results of the extraction of verbal chunks. It may be summarized by the following question: what kind and what quantity of new information does one need to significantly increase the results ? 

Besides this, we also plan to:
\begin{itemize}
\item extend the system and develop our set of linguistic rules to perform the extraction of all verbal forms, including gerund and present participle chunks,

\item further describe the potential incidental clauses whithin verbal chunks,

\item extend the coverage of the system to verbal coordination of finite and infinitive verb, as well as participial forms in French,

\item take advantage of collateral benefits of the 
system: if verbal chunks are correctly specified, we may incidentally deduce that all occurrences of ambiguous forms not in verbal chunks must not be associated to the function or the category to which they are associated in verbal chunks.
Eg., all {\it lui} forms not in verbal chunks and not following a 
preposition must be associated to a nominative role~\footnote{In B\`es \& Dahl \cite{bes-03}, verbal chunks
along with phrase introducers (e.g. complementizers, interrogative expressions
or conjunctions) are crucial for the obtention of coordination of verbal phrases.}.
\end{itemize}
Future work will also consists in testing the system on corpora from various sources and of different types, the goal being to isolate different sets of rules with respect to their precision and effectiveness in terms of recall. This next system could then become a modular system which the user might use with higher precision/lower recall or lower precision/higher recall, depending on his needs.  

In other respects, we are also focused on the applicative dimension of a such system, and one application domain in which one would need high precision is teaching.
In the field of language acquisition, our system may be used to set up a large variety of exercises using local contexts. The rule-based system, provided it is reliable, and the evaluation system can also be used in order to automatically detect the errors made by the students in the exercises. By integrating this system in a specific software, we are searching two goals: one is to link development in Natural Language Processing with the issue of French grammar teaching to foreign students and the other is to share the different concerns of NLP and teaching methods.

%
%

\end{document}